\documentclass[letterpaper,10pt,conference]{ieeeconf}
\IEEEoverridecommandlockouts
\usepackage{amsmath} 
\usepackage{amssymb}  
\usepackage{fix-cm}
\usepackage{pifont}
\usepackage{xcolor}
\usepackage{graphicx}
\usepackage{booktabs}
\usepackage{tabularx}
\usepackage{microtype}
\usepackage{cite}
\usepackage{multirow}
\usepackage{needspace}
\usepackage{url}
\usepackage{etoolbox}
\makeatletter
\patchcmd{\@maketitle}{\vskip0.25in}{\vskip0.10in}{}{\errmessage{Title spacing patch failed}}
\makeatother

\makeatletter
\newcommand{\CoordiBibControl}{%
  \immediate\write\@auxout{\string\citation{CoordiBibControl}}%
}
\makeatother

\makeatletter
\newcommand{\widefigurecaption}[1]{%
  \refstepcounter{figure}%
  \@makecaption{\fnum@figure}{#1}%
}
\makeatother

\newcommand{\cmark}{\textcolor{green!50!black}{\ding{51}}}
\newcommand{\xmark}{\textcolor{red!75!black}{\ding{55}}}
\title{\LARGE \bf
Coordi-$\pi$: Coordination-Aware Dual-Hand Action Representation for Bimanual Dexterous Manipulation
}

\author{Yiming Meng$^{1,*}$, Yanming Shao$^{2,*}$, Weijie Shen$^{1}$, Kunyang Lin$^{2}$,\\
Jiayu Sun$^{1}$, Ping Luo$^{2,\dagger}$, Yao Mu$^{1,\dagger}$\\[1pt]
{\normalsize $^{1}$Shanghai Jiao Tong University, $^{2}$The University of Hong Kong}\\[0pt]
{\normalsize $^{*}$Equal contribution. \quad $^{\dagger}$Corresponding authors.}
}

\begin{document}
\CoordiBibControl

\newcommand{\FirstPageTeaser}{%
  \begin{minipage}{\textwidth}
    \centering
    \includegraphics[
      width=0.92\linewidth,
      trim=0 22bp 0 12bp,clip 
    ]{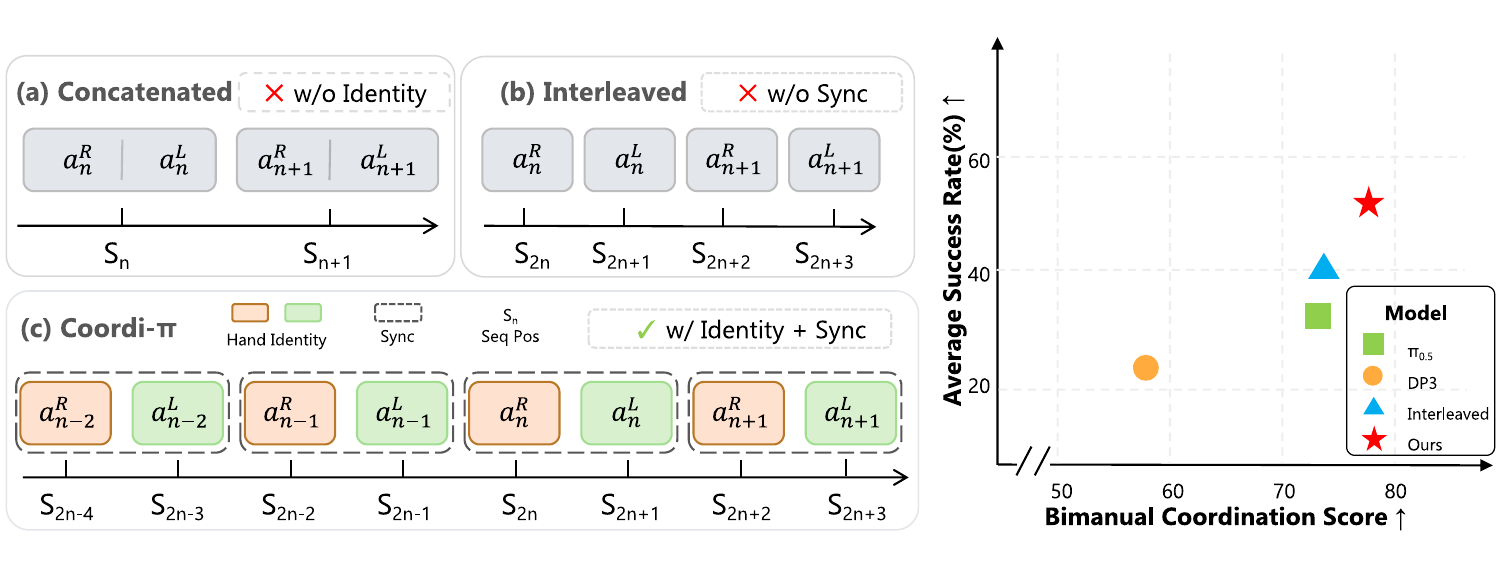}

    \widefigurecaption{
    \textbf{Action representations and coordination performance.}
    (a) Concatenation packs both arm--hand actions into one token per
    control timestep.
    (b) Interleaving separates the two sides, but token adjacency alone
    does not distinguish same-step from cross-step actions.
    (c) Coordi-$\pi$ explicitly encodes hand identity and control timestep.
    Orange and green indicate right- and left-side actions; dashed boxes
    group actions sharing a control timestep.
    \textbf{Right:} Average success rate and Bimanual Coordination Score
    (BCS) on the five bimanual DexJoCo tasks under \textit{rand-obj};
    higher is better for both metrics.
    }
    \label{fig:action_representation}

    \vspace{5pt}
  \end{minipage}%
}

\IEEEaftertitletext{\vspace{-14pt}\FirstPageTeaser}

\maketitle
\thispagestyle{empty}
\pagestyle{empty}

\begin{abstract}
\looseness=-1
Vision-language-action (VLA) models have shown strong potential for general-purpose robotic manipulation, but extending them to bimanual dexterous manipulation introduces a high-dimensional and structured action space. Existing bimanual VLA policies commonly concatenate right- and left-side arm--hand actions at each control timestep, leaving cross-hand interactions implicit within a single action vector. We introduce Coordi-$\pi$, a structured action representation for bimanual dexterous manipulation. Coordi-$\pi$ represents the right- and left-side arm--hand actions as individual action tokens and organizes them into an interleaved sequence, explicitly exposing the two sides within the action sequence. To explicitly encode hand identity and temporal synchronization in this interleaved representation, a learnable hand-identity embedding distinguishes the two sides, while sequence--control-timestep factorized rotary positional encoding (RoPE) separates sequence position from control timestep, allowing actions at the same control timestep to share a temporal coordinate while retaining distinct sequence positions. Across all five bimanual dexterous manipulation tasks in DexJoCo, Coordi-$\pi$ achieves state-of-the-art performance under both \textit{rand-obj} and \textit{rand-full}, improving the average success rate over $\pi_{0.5}$ from 31.9\% to 51.3\% and from 18.3\% to 32.3\%, respectively. Coordi-$\pi$ also consistently outperforms $\pi_{0.5}$ across four real-world bimanual dexterous manipulation tasks.
\end{abstract}

\section{INTRODUCTION}

\looseness=-1
Vision-language-action (VLA) models have recently shown strong progress
in general-purpose robotic manipulation~\cite{rt1,rt2,openx,octo,openvla,pi0,pi05}. While early generalist policies largely focused on single-arm manipulation with parallel-jaw grippers, recent work has extended VLA-based control to higher-DoF embodiments, including dexterous hands, humanoid robots, and dual-arm systems~\cite{hpt,gr00tn1,dexvla,dexora,dexjoco}. Bimanual dexterous manipulation further requires each side to control a high-DoF arm--hand system while coordinating actions across
the two sides~\cite{buds}, substantially increasing both the
dimensionality and structural complexity of the joint action space.

Despite recent progress, action representation remains underexplored
for bimanual dexterous manipulation. Conventional approaches typically
concatenate the right- and left-side arm--hand actions at each control
timestep,
\begin{equation}
\mathbf{a}_t =
[\mathbf{a}_t^R;\mathbf{a}_t^L],
\end{equation}
where $\mathbf{a}_t^R$ and $\mathbf{a}_t^L$ denote the corresponding arm--hand actions. Although this representation naturally preserves the joint action executed at each timestep, the two sides are distinguished only by the feature dimensions they occupy, leaving their identities and cross-hand structure implicit within a single action vector. In contrast, representing the two sides as individual action tokens in an interleaved sequence,
\begin{equation}
R_0,L_0,R_1,L_1,\ldots,R_{H-1},L_{H-1},
\end{equation}
makes the bimanual structure explicit and enables direct token-level interaction. However, interleaving breaks the one-to-one correspondence between
sequence position and control timestep: adjacent token pairs may
represent either synchronized or temporally consecutive actions.

To address these representation issues, we introduce \textbf{Coordi-$\pi$}, a structured action representation framework for bimanual dexterous manipulation, as illustrated in Fig.~\ref{fig:action_representation}. Coordi-$\pi$ first represents each bimanual action chunk as an
interleaved sequence of right- and left-side arm--hand tokens. A learnable hand-identity embedding distinguishes the two sides within the shared action representation. We further introduce sequence--control-timestep factorized rotary positional encoding (RoPE), which assigns each action token both a sequence-position coordinate and a control-timestep coordinate. In this representation, $R_h$ and $L_h$ occupy different positions
in the token sequence while sharing the same control-timestep
coordinate, explicitly representing their temporal synchronization.

We evaluate Coordi-$\pi$ on five bimanual dexterous manipulation
tasks in DexJoCo~\cite{dexjoco} under both \textit{rand-obj} and
the more challenging \textit{rand-full} setting, and further evaluate
it on four real-world bimanual dexterous manipulation tasks.
Coordi-$\pi$ achieves state-of-the-art performance on DexJoCo,
improving the average success rate over $\pi_{0.5}$ from $31.9\%$
to $51.3\%$ under \textit{rand-obj} and from $18.3\%$ to $32.3\%$
under \textit{rand-full}. Coordi-$\pi$ also outperforms $\pi_{0.5}$
across all four real-world tasks. Ablation studies further show
complementary contributions from the interleaved action representation,
hand-identity embedding, and sequence--control-timestep factorized RoPE.

In summary, the contributions of this paper are threefold:
\begin{itemize}

\item We propose Coordi-$\pi$, a structured action representation for bimanual dexterous manipulation that represents right- and left-side arm--hand actions as individual tokens and organizes them in an interleaved sequence, explicitly exposing the two sides within the action sequence.

\item We introduce a learnable hand-identity embedding and sequence--control-timestep factorized RoPE to represent hand identity and temporal synchronization in the interleaved action representation.

\item We show that Coordi-$\pi$ achieves state-of-the-art performance
across all five bimanual dexterous manipulation tasks in DexJoCo under both
\textit{rand-obj} and \textit{rand-full}, and outperforms $\pi_{0.5}$
across all four real-world tasks.

\end{itemize}

\section{RELATED WORK}

\begin{figure*}[t]
    \centering
    \includegraphics[width=0.90\textwidth,trim=0 1bp 0 12bp,clip]{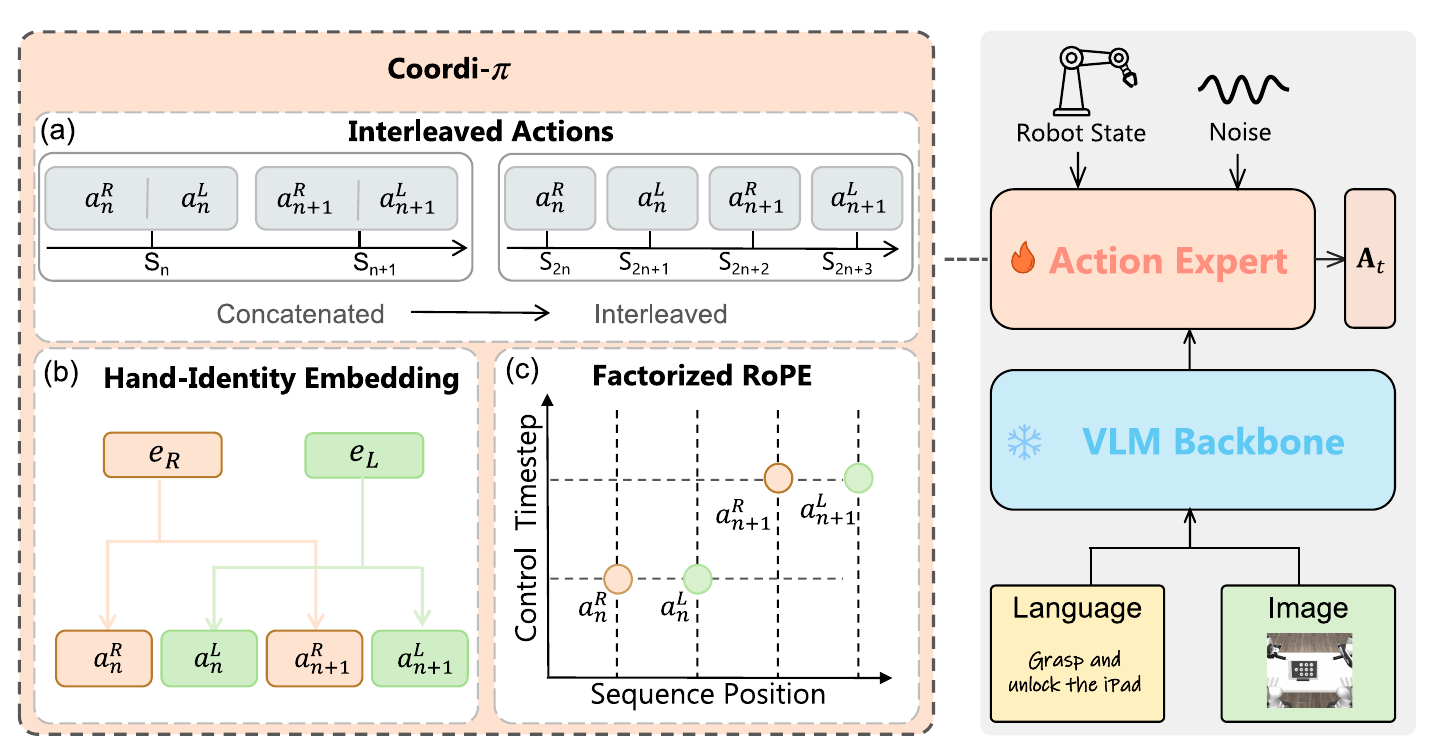}
    \caption{
    \textbf{Overview of Coordi-$\pi$.}
    Within the action expert, (a) joint bimanual actions are split into
    interleaved right- and left-side tokens, and (b) learnable
    hand-identity embeddings distinguish the two sides.
    (c) Factorized RoPE encodes sequence position and control timestep
    separately: paired actions occupy different sequence positions but
    share the same control-timestep coordinate.
    \textbf{Right:} Conditioned on images, language, and robot state,
    the action expert generates an action chunk from noise.
    Output tokens are paired by control timestep into joint
    bimanual commands, without changing the action horizon.
    }
    \label{fig:overview}
\end{figure*}

\textbf{Bimanual and Dexterous Manipulation.}
\looseness=-1
Recent work has advanced bimanual dexterous manipulation through
improved teleoperation, data collection, policy learning, and
evaluation. BiDex develops a high-DoF teleoperation system for
collecting bimanual dexterous demonstrations~\cite{bidex}, while
DexMimicGen generates large-scale demonstration data from a small
number of human demonstrations~\cite{dexmimicgen}. Related systems
such as Mobile ALOHA~\cite{mobilealoha}, ALOHA
Unleashed~\cite{alohaunleashed}, and DexCap~\cite{dexcap} further
explore scalable data collection and policy learning for complex
bimanual or dexterous manipulation. BiDexHands studies high-DoF
bimanual dexterous control in simulation~\cite{bidexhands}.
On the policy side, InterACT captures inter-dependencies between two
arms during action prediction~\cite{interact}, and KStar Diffuser
incorporates spatial--temporal robot structure and kinematic
constraints into bimanual control~\cite{kstar}. RDT-1B develops a
diffusion foundation model for bimanual manipulation~\cite{rdt1b},
while Dexora targets high-DoF dual-arm dual-hand VLA
control~\cite{dexora}. DexJoCo provides a benchmark covering bimanual
coordination, tool use, and long-horizon dexterous
manipulation~\cite{dexjoco}. Coordi-$\pi$ focuses specifically on action representation for bimanual dexterous VLA control, where each side controls a
high-DoF arm--hand system.

\textbf{Vision-Language-Action Policies.}
\looseness=-1
Vision-language-action policies leverage large-scale visual,
language, and robot data for general-purpose robot control.
PaLM-E~\cite{palme} and VIMA~\cite{vima} demonstrate early forms of
multimodal reasoning and prompting for embodied manipulation, while
RT-2 transfers vision-language knowledge directly to robotic
actions~\cite{rt2}. OpenVLA provides an open-source VLA pretrained on diverse robot demonstrations~\cite{openvla}, alongside the broader development of large-scale robot datasets such as DROID~\cite{droid}.
For continuous action generation, $\pi_0$ combines a pretrained
vision-language backbone with a flow-matching action
expert~\cite{pi0}, and $\pi_{0.5}$ further extends this framework
toward open-world manipulation~\cite{pi05}. Recent work has also
extended VLA policies to more diverse robot embodiments. DexVLA
studies general robot control across single-arm, bimanual, and
dexterous-hand settings through a diffusion-based action
expert~\cite{dexvla}. Coordi-$\pi$ builds on this line of VLA policies
but focuses on the action representation required for bimanual
dexterous control.

\textbf{Action Representation and Sequence Modeling.}
\looseness=-1
Robot policies represent continuous actions in different forms. ACT predicts temporally extended action chunks~\cite{act}, while Diffusion Policy represents continuous action trajectories through conditional diffusion~\cite{diffusionpolicy}. Within VLA policies, OpenVLA-OFT adopts continuous action prediction together with parallel decoding and action chunking~\cite{openvlaoft}, while FAST and ActionCodec study tokenization of continuous robot actions~\cite{fast,actioncodec}. SpatialVLA introduces spatially structured representations for robot actions~\cite{spatialvla}, and Ego-Pi adopts an interleaved action formulation for high-dimensional dexterous action prediction within a $\pi_{0.5}$-based policy~\cite{egopi}. Positional structure in sequence models is commonly represented with RoPE~\cite{rope}, while multimodal extensions such as M-RoPE and VideoRoPE factor positional coordinates across spatial and temporal dimensions~\cite{qwen2vl,videorope}. Coordi-$\pi$ focuses on structured action representation for bimanual control, combining interleaved arm--hand tokens with explicit hand identity and control-timestep structure.

\section{METHOD}

\begingroup
\setlength{\abovedisplayskip}{5pt plus 2pt minus 1pt}
\setlength{\belowdisplayskip}{5pt plus 2pt minus 1pt}
\setlength{\abovedisplayshortskip}{0pt plus 2pt}
\setlength{\belowdisplayshortskip}{5pt plus 2pt minus 1pt}

\subsection{Overview}

Coordi-$\pi$ is built on the $\pi_{0.5}$ VLA architecture~\cite{pi05}
and modifies the internal action representation for bimanual dexterous
control. As illustrated in Fig.~\ref{fig:overview}, Coordi-$\pi$
represents right- and left-side arm--hand actions as an interleaved
token sequence, augments each token with a learnable hand-identity
embedding, and applies sequence--control-timestep factorized RoPE to
represent temporal synchronization.

\subsection{Problem Formulation}

At control timestep $t$, the policy receives an observation
$\mathbf{o}_t=(\mathbf{I}_t,\mathbf{l},\mathbf{s}_t)$,
where $\mathbf{I}_t$ denotes the visual observations, $\mathbf{l}$ the
language instruction, and $\mathbf{s}_t$ the robot proprioceptive state.
The policy predicts an $H$-step bimanual action chunk
\begin{equation}
\mathbf{A}_t
=
\left[
\mathbf{a}_t,
\mathbf{a}_{t+1},
\ldots,
\mathbf{a}_{t+H-1}
\right].
\end{equation}
Each bimanual action consists of right- and left-side arm--hand commands,
\begin{equation}
\mathbf{a}_{t+h}
=
\left[
\mathbf{a}_{t+h}^{R};
\mathbf{a}_{t+h}^{L}
\right],
\qquad
\mathbf{a}_{t+h}^{R},
\mathbf{a}_{t+h}^{L}
\in
\mathbb{R}^{d},
\end{equation}
where $h=0,\ldots,H-1$ and $d$ denotes the per-side action dimension.
Under the conventional bimanual action representation, the two sides
are represented within different feature dimensions of the same action
vector.

\subsection{Interleaved Bimanual Action Representation}

Coordi-$\pi$ moves the bimanual structure from the feature dimension to
the token dimension by assigning the right- and left-side actions to
separate action tokens. For $h=0,\ldots,H-1$, we define
\begin{equation}
\mathbf{z}_{2h}
=
\mathbf{a}_{t+h}^{R},
\qquad
\mathbf{z}_{2h+1}
=
\mathbf{a}_{t+h}^{L}.
\end{equation}
The resulting sequence
$\mathbf{Z}_t=[\mathbf{z}_0,\mathbf{z}_1,\ldots,\mathbf{z}_{2H-1}]$
corresponds to
\begin{equation}
R_0,L_0,R_1,L_1,\ldots,R_{H-1},L_{H-1},
\end{equation}
where $R_h$ and $L_h$ denote the right- and left-side actions at control
timestep $t+h$. Thus, an $H$-step bimanual action chunk is represented
by $2H$ side-specific action tokens. The $2H$ tokens still correspond to an $H$-step action chunk, with
$R_h$ and $L_h$ paired as the joint action at timestep $t+h$.

The interleaved action representation makes the two arm--hand systems
explicit in the token sequence and places the synchronized actions
$R_h$ and $L_h$ in neighboring token positions. However, interleaving
alone does not explicitly represent hand identity, and sequence position
no longer coincides with control timestep. We address these two aspects
with a learnable hand-identity embedding and sequence--control-timestep
factorized RoPE, respectively.

\subsection{Hand-Identity Embedding}

Right- and left-side action tokens share the same action format and input
projection. They also share the same output projection, while hand identity is
provided by the learnable embedding. We introduce a learnable hand-identity embedding to explicitly
represent which side each action token belongs to. Let
$f_{\mathrm{in}}(\cdot)$ denote the shared action projection. The input
representation of action token $i$ is
\begin{equation}
\begin{aligned}
\mathbf{h}_i &= f_{\mathrm{in}}(\mathbf{z}_i)+\mathbf{e}_{h(i)},\\
h(i) &= \begin{cases}
R, & i \bmod 2 = 0,\\
L, & i \bmod 2 = 1,
\end{cases}
\end{aligned}
\end{equation}
and $\mathbf{e}_{R}$ and $\mathbf{e}_{L}$ are learnable hand-identity
embeddings for the right and left sides, respectively. The same embedding
is shared by all action tokens from the same side across the action chunk.
Both embeddings are initialized to zero.

\begin{figure*}[t]
    \centering
    \includegraphics[width=\textwidth]
    {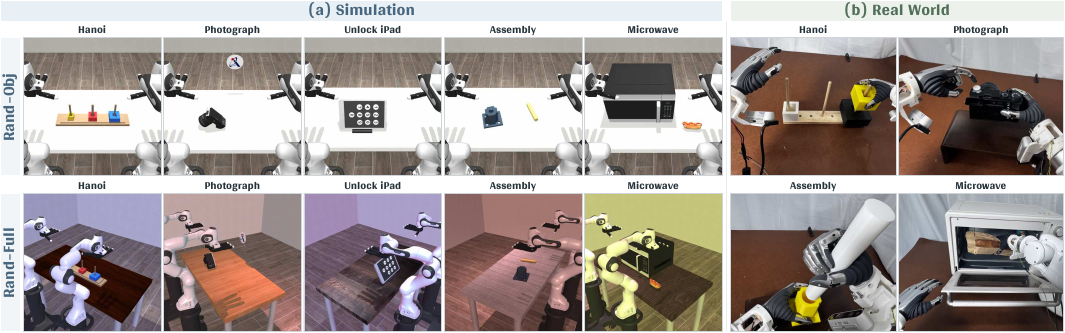}
    \caption{Overview of the evaluation tasks.
    (a) All five bimanual dexterous manipulation tasks in DexJoCo under
    \textit{rand-obj} and \textit{rand-full}.
    (b) Four real-world bimanual dexterous manipulation tasks.}
    \label{fig:task_overview}
\end{figure*}

\subsection{Sequence--Control-Timestep Factorized RoPE}

The interleaved action representation introduces two different notions
of position: sequence position and control timestep. For action token
$i$, we define the corresponding coordinates as
\begin{equation}
s_i=i,\qquad \tau_i=\left\lfloor\frac{i}{2}\right\rfloor.
\end{equation}
For an $H$-step action chunk, the corresponding coordinate sequences are
\begin{equation}
\begin{aligned}
\mathbf{s}&=[0,1,2,3,\ldots,2H-2,2H-1],\\
\boldsymbol{\tau}&=[0,0,1,1,\ldots,H-1,H-1].
\end{aligned}
\end{equation}

For the synchronized pair $R_h$ and $L_h$,
\begin{equation}
\begin{aligned}
s_{R_h}&=2h,\qquad s_{L_h}=2h+1,\\
\tau_{R_h}&=\tau_{L_h}=h.
\end{aligned}
\end{equation}
Thus, $R_h$ and $L_h$ occupy different sequence positions while sharing
the same control timestep.

An important distinction arises between token adjacency and control
time. In the interleaved sequence, $R_h \rightarrow L_h$ and
$L_h \rightarrow R_{h+1}$ are both adjacent token pairs and therefore
have the same sequence-position offset, $\Delta s=1$. Their temporal
relations, however, are different: $R_h$ and $L_h$ belong to the same
control timestep, giving $\Delta\tau=0$, whereas $L_h$ and
$R_{h+1}$ span consecutive timesteps, giving $\Delta\tau=1$.

A sequence-position RoPE represents both pairs with the same relative
sequence offset and does not explicitly encode this difference in
control time. We therefore factorize RoPE into a sequence-position
subspace and a control-timestep subspace of equal size,
$d_{\mathrm{seq}}=d_{\mathrm{ctrl}}=d_{\mathrm{head}}/2$.

The query and key vectors are partitioned accordingly,
\begin{equation}
\mathbf{q}_i
=
\left[
\mathbf{q}_i^{\mathrm{seq}};
\mathbf{q}_i^{\mathrm{ctrl}}
\right],
\qquad
\mathbf{k}_i
=
\left[
\mathbf{k}_i^{\mathrm{seq}};
\mathbf{k}_i^{\mathrm{ctrl}}
\right].
\end{equation}
RoPE is applied independently to the two subspaces using their
corresponding coordinates,
\begin{equation}
\begin{aligned}
\tilde{\mathbf{q}}_i
&=
\left[
\operatorname{RoPE}
\left(
\mathbf{q}_i^{\mathrm{seq}},
s_i
\right);
\operatorname{RoPE}
\left(
\mathbf{q}_i^{\mathrm{ctrl}},
\tau_i
\right)
\right],\\
\tilde{\mathbf{k}}_i
&=
\left[
\operatorname{RoPE}
\left(
\mathbf{k}_i^{\mathrm{seq}},
s_i
\right);
\operatorname{RoPE}
\left(
\mathbf{k}_i^{\mathrm{ctrl}},
\tau_i
\right)
\right].
\end{aligned}
\end{equation}

\looseness=-1
Factorized RoPE is used to compute attention scores between action
tokens; attention involving context tokens retains the original
positional encoding.
For the synchronized pair $(R_h,L_h)$, $\Delta s=1$ and
$\Delta\tau=0$.
The sequence-position coordinate preserves the ordering of the
interleaved token sequence, while the shared control-timestep coordinate
represents temporal synchronization between the two sides.
Interleaving only changes the internal action representation; the
right- and left-side actions associated with the same control timestep
are decoded back into a joint bimanual command for execution.

\subsection{Training and Action Decoding}

Coordi-$\pi$ retains the flow-matching objective of
$\pi_{0.5}$~\cite{pi0,pi05}. During training, noisy bimanual action
chunks are arranged in the interleaved token order, with hand-identity
embeddings added to the corresponding action tokens and factorized RoPE
applied using their sequence-position and control-timestep coordinates.
The control-timestep coordinate indexes the temporal structure within
the action chunk and is independent of the continuous flow-matching
time.

At inference, the generated side-specific action tokens are
de-interleaved and paired according to their shared control timestep.
For each timestep $t+h$, the bimanual action is reconstructed as
\begin{equation}
\hat{\mathbf{a}}_{t+h}
=
\left[
\hat{\mathbf{a}}_{t+h}^{R};
\hat{\mathbf{a}}_{t+h}^{L}
\right],
\qquad
h=0,\ldots,H-1.
\end{equation}

\endgroup

\begin{table*}[t]
\centering
\caption{Success rates (\%) on all five bimanual tasks in DexJoCo. \textnormal{Obj} and
\textnormal{Full} denote \textnormal{\textit{rand-obj}} and \textnormal{\textit{rand-full}},
respectively. Entries are mean $\pm$ standard deviation; best results
per task and setting are bold.}
\label{tab:main_results}
\footnotesize
\setlength{\tabcolsep}{1.1pt}
\medmuskip=1.5mu
\renewcommand{\arraystretch}{1.05}
\begin{tabular*}{\textwidth}{@{\extracolsep{\fill}}l*{12}{c}@{}}
\toprule
\multirow{2}{*}{Method}
& \multicolumn{2}{c}{Hanoi}
& \multicolumn{2}{c}{Photograph}
& \multicolumn{2}{c}{Unlock iPad}
& \multicolumn{2}{c}{Assembly}
& \multicolumn{2}{c}{Microwave}
& \multicolumn{2}{c}{Avg.} \\
\cmidrule(lr){2-3}\cmidrule(lr){4-5}\cmidrule(lr){6-7}
\cmidrule(lr){8-9}\cmidrule(lr){10-11}\cmidrule(l){12-13}
& Obj & Full & Obj & Full & Obj & Full
& Obj & Full & Obj & Full & Obj & Full \\
\midrule

ACT~\cite{act}
& $6.0 \pm 2.0$
& $4.7 \pm 2.3$
& $7.3 \pm 1.2$
& $3.3 \pm 1.2$
& $9.3 \pm 3.1$
& $0.7 \pm 1.2$
& $0.0 \pm 0.0$
& $0.0 \pm 0.0$
& $66.0 \pm 2.0$
& $50.0 \pm 6.9$
& $17.7$
& $11.7$ \\

DP-T~\cite{diffusionpolicy}
& $24.7 \pm 4.6$
& $0.7 \pm 1.2$
& $56.7 \pm 4.6$
& $7.3 \pm 1.2$
& $8.0 \pm 2.0$
& $2.0 \pm 2.0$
& $4.7 \pm 3.1$
& $0.0 \pm 0.0$
& $73.3 \pm 11.6$
& $21.3 \pm 4.6$
& $33.5$
& $6.3$ \\

DP-C~\cite{diffusionpolicy}
& $12.7 \pm 3.1$
& $9.3 \pm 6.1$
& $24.0 \pm 8.7$
& $8.7 \pm 4.2$
& $\mathbf{52.0 \pm 2.0}$
& $12.0 \pm 3.5$
& $3.3 \pm 1.2$
& $0.0 \pm 0.0$
& $54.0 \pm 12.5$
& $62.7 \pm 6.4$
& $29.2$
& $18.5$ \\

DP3~\cite{dp3}
& $26.0 \pm 2.0$
& $13.3 \pm 4.2$
& $24.0 \pm 9.2$
& $13.3 \pm 9.5$
& $16.0 \pm 4.0$
& $13.3 \pm 1.2$
& $4.7 \pm 1.2$
& $0.0 \pm 0.0$
& $45.3 \pm 16.2$
& $36.7 \pm 6.1$
& $23.2$
& $15.3$ \\

GR00T N1.5~\cite{gr00tn15}
& $0.7 \pm 1.2$
& $0.0 \pm 0.0$
& $40.7 \pm 7.0$
& $18.7 \pm 7.0$
& $12.7 \pm 11.0$
& $0.0 \pm 0.0$
& $0.7 \pm 1.2$
& $1.3 \pm 1.2$
& $50.7 \pm 4.6$
& $42.0 \pm 7.2$
& $21.1$
& $12.4$ \\

$\pi_{0.5}$~\cite{pi05}
& $15.3 \pm 3.1$
& $15.3 \pm 2.3$
& $56.7 \pm 5.0$
& $21.3 \pm 2.3$
& $12.0 \pm 3.5$
& $0.0 \pm 0.0$
& $5.3 \pm 1.2$
& $0.0 \pm 0.0$
& $70.0 \pm 3.5$
& $54.7 \pm 6.1$
& $31.9$
& $18.3$ \\

\midrule

\textbf{Coordi-$\pi$}
& $\mathbf{48.7 \pm 1.2}$
& $\mathbf{36.0 \pm 2.0}$
& $\mathbf{72.0 \pm 6.0}$
& $\mathbf{28.7 \pm 9.9}$
& $\mathbf{52.0 \pm 3.5}$
& $\mathbf{28.0 \pm 4.0}$
& $\mathbf{7.3 \pm 1.2}$
& $\mathbf{2.0 \pm 2.0}$
& $\mathbf{76.7 \pm 8.3}$
& $\mathbf{66.7 \pm 11.7}$
& $\mathbf{51.3}$
& $\mathbf{32.3}$ \\

\bottomrule
\end{tabular*}
\end{table*}

\section{EXPERIMENTS}

We evaluate Coordi-$\pi$ along four aspects:
(1) \textbf{Performance:} whether the proposed bimanual action
representation improves task success over existing policies on
dual-arm dexterous manipulation;
(2) \textbf{Ablations:} the contributions of the hand-identity
embedding and sequence--control-timestep factorized RoPE beyond the
interleaved action representation;
(3) \textbf{Coordination:} whether the performance gains are accompanied
by improved bimanual coordination beyond task completion; and
(4) \textbf{Robustness and Real-World Performance:} whether the
advantages persist under stronger environment randomization and on
a real-world dual-arm dexterous robotic system.

\subsection{Experimental Setup}

\textbf{Tasks and Randomization.}
We evaluate Coordi-$\pi$ on all five bimanual dexterous manipulation
tasks in DexJoCo~\cite{dexjoco}: \textit{Hanoi}, \textit{Photograph},
\textit{Unlock iPad}, \textit{Assembly}, and \textit{Microwave}.
The tasks cover complementary bimanual roles, sequential interactions,
precise insertion, and long-horizon manipulation. We consider both
\textit{rand-obj} and \textit{rand-full}. The former randomizes object
placement and table height, while the latter additionally varies
third-person camera poses, illumination, and tabletop textures.
We further evaluate \textit{Photograph}, \textit{Hanoi},
\textit{Assembly}, and \textit{Microwave} on a real-world dual-arm
dexterous platform. Figure~\ref{fig:task_overview} summarizes the
simulation and real-world evaluation tasks.

\textbf{Observation and Action Space.}
The policy receives three $224\times224$ RGB views, a language
instruction, and a 46-D proprioceptive state. We use an action horizon
of $H=30$. Each side has a 22-D arm--hand action, yielding a 44-D
bimanual command at each control timestep. Under Coordi-$\pi$, the same
$H$-step action chunk is represented internally as $2H=60$
side-specific action tokens.

\textbf{Baselines.}
We compare Coordi-$\pi$ with ACT~\cite{act}, DP-T and
DP-C~\cite{diffusionpolicy}, DP3~\cite{dp3}, GR00T
N1.5~\cite{gr00tn15}, and $\pi_{0.5}$~\cite{pi05}.
We report the DexJoCo benchmark results for ACT, DP-T, DP-C,
GR00T N1.5, and $\pi_{0.5}$, and evaluate DP3 under the same protocol.
Coordi-$\pi$ and all ablation variants use $\pi_{0.5}$ as the base VLA.

\textbf{Training and Evaluation.}
\looseness=-1
Following the standard DexJoCo protocol, Coordi-$\pi$ and all ablation
variants are trained for 40k iterations with a batch size of 32 on four
A100 GPUs using the fixed training seed of 42.
For our simulation evaluations, each method--task--setting combination
is evaluated using three evaluation seeds (0, 1, and 2), with
50 episodes per seed, resulting in 150 evaluation episodes per
combination. We report the mean success rate across the three
evaluation seeds, with $\pm$ denoting the sample standard deviation.
Results taken from DexJoCo are reported as published.
Unless otherwise specified, all ablation variants use the same
training data, training budget, and evaluation protocol.

\subsection{Main Results on DexJoCo}

\looseness=-1
Under \textit{rand-obj}, Coordi-$\pi$ achieves the highest average
success rate across all five bimanual dexterous manipulation tasks,
reaching $51.3\%$, compared with $33.5\%$ for DP-T and $31.9\%$ for
$\pi_{0.5}$. This corresponds to a $19.4$ percentage-point improvement
over $\pi_{0.5}$. Coordi-$\pi$ achieves the best or tied-best
performance on all five tasks.

\looseness=-1
The largest gains are observed on \textit{Hanoi} and
\textit{Unlock iPad}, where Coordi-$\pi$ improves over $\pi_{0.5}$
from $15.3\%$ to $48.7\%$ and from $12.0\%$ to $52.0\%$,
respectively. Both tasks require complementary hand roles and
ordered cross-hand interactions.

\looseness=-1
On \textit{Microwave}, Coordi-$\pi$ still improves over $\pi_{0.5}$,
although by a smaller margin than on \textit{Hanoi} and
\textit{Unlock iPad}. \textit{Assembly} remains difficult for all
methods, with Coordi-$\pi$ reaching only $7.3\%$. The low success rate
indicates that precise contact and insertion remain major bottlenecks
beyond the choice of bimanual action representation.

\Needspace{4\baselineskip}
\subsection{Ablation Study}
\label{sec:ablation}

We study the contributions of the hand-identity embedding and
sequence--control-timestep factorized RoPE on top of the interleaved
action representation. All variants use the same training data,
optimization budget, and evaluation protocol.

\begin{table}[!t]
\centering
\caption{Ablation success rates (\%) under \textnormal{\textit{rand-obj}} on all five
bimanual dexterous manipulation tasks in DexJoCo. All variants use the
interleaved action representation. Best results are shown in bold.}
\label{tab:ablation}
\footnotesize
\setlength{\tabcolsep}{1.5pt}
\renewcommand{\arraystretch}{1.12}
\begin{tabular*}{\columnwidth}{@{\extracolsep{\fill}}lcccc@{}}
\toprule
& \shortstack{Inter-\\leaved} & + Hand ID & + RoPE & \textbf{Coordi-$\pi$} \\
\midrule
Hand ID & \xmark & \cmark & \xmark & \cmark \\
Factorized RoPE & \xmark & \xmark & \cmark & \cmark \\
\midrule
Hanoi & $37.3\pm9.9$ & $43.3\pm7.0$ & $39.3\pm4.6$ & $\mathbf{48.7\pm1.2}$ \\
Photograph & $58.0\pm5.3$ & $69.3\pm6.4$ & $67.3\pm3.1$ & $\mathbf{72.0\pm6.0}$ \\
Unlock iPad & $29.3\pm7.0$ & $45.3\pm13.3$ & $49.3\pm5.8$ & $\mathbf{52.0\pm3.5}$ \\
Assembly & $4.7\pm1.2$ & $3.3\pm2.3$ & $2.7\pm1.2$ & $\mathbf{7.3\pm1.2}$ \\
Microwave & $69.3\pm7.0$ & $50.7\pm7.0$ & $62.7\pm5.0$ & $\mathbf{76.7\pm8.3}$ \\
\midrule
Avg. & $39.7$ & $42.4$ & $44.3$ & $\mathbf{51.3}$ \\
\bottomrule
\end{tabular*}
\end{table}

\begin{table*}[t]
\centering
\caption{Bimanual coordination scores on all five bimanual dexterous
manipulation tasks in DexJoCo under \textnormal{\textit{rand-obj}}. RCS, TCS, and BCS range from 0 to 100;
higher is better.}
\label{tab:coordination}
\footnotesize
\setlength{\tabcolsep}{3.4pt}
\renewcommand{\arraystretch}{1.08}
\begin{tabular*}{\textwidth}{@{\extracolsep{\fill}}lcccccccccccc@{}}
\toprule
& \multicolumn{3}{c}{$\pi_{0.5}$}
& \multicolumn{3}{c}{DP3}
& \multicolumn{3}{c}{Interleaved}
& \multicolumn{3}{c}{\textbf{Coordi-$\pi$}} \\
\cmidrule(lr){2-4}\cmidrule(lr){5-7}
\cmidrule(lr){8-10}\cmidrule(lr){11-13}
Task & RCS & TCS & BCS & RCS & TCS & BCS
& RCS & TCS & BCS & RCS & TCS & BCS \\
\midrule
Hanoi
& 91.00 & 73.77 & 82.39
& 89.17 & 67.49 & 78.33
& \textbf{96.00} & \textbf{83.50} & \textbf{89.75}
& 93.00 & 79.65 & 86.32 \\
Photograph
& 68.79 & 68.52 & 68.65
& 41.58 & 30.46 & 36.02
& 71.31 & 70.33 & 70.82
& \textbf{72.58} & \textbf{76.57} & \textbf{74.58} \\
Unlock iPad
& 77.91 & 62.17 & 70.04
& 64.05 & 35.68 & 49.86
& 75.10 & 58.96 & 67.03
& \textbf{81.19} & \textbf{70.40} & \textbf{75.80} \\
Assembly
& 80.56 & 35.31 & 57.94
& 76.17 & 22.30 & 49.23
& 72.57 & 30.16 & 51.37
& \textbf{81.22} & \textbf{41.57} & \textbf{61.39} \\
Microwave
& 98.00 & 67.21 & 82.61
& 97.93 & 58.85 & 78.39
& 99.33 & 75.54 & 87.44
& \textbf{100.00} & \textbf{82.43} & \textbf{91.21} \\
\bottomrule
\end{tabular*}
\end{table*}

\begin{figure}[t]
    \centering
    \includegraphics[width=0.88\columnwidth]{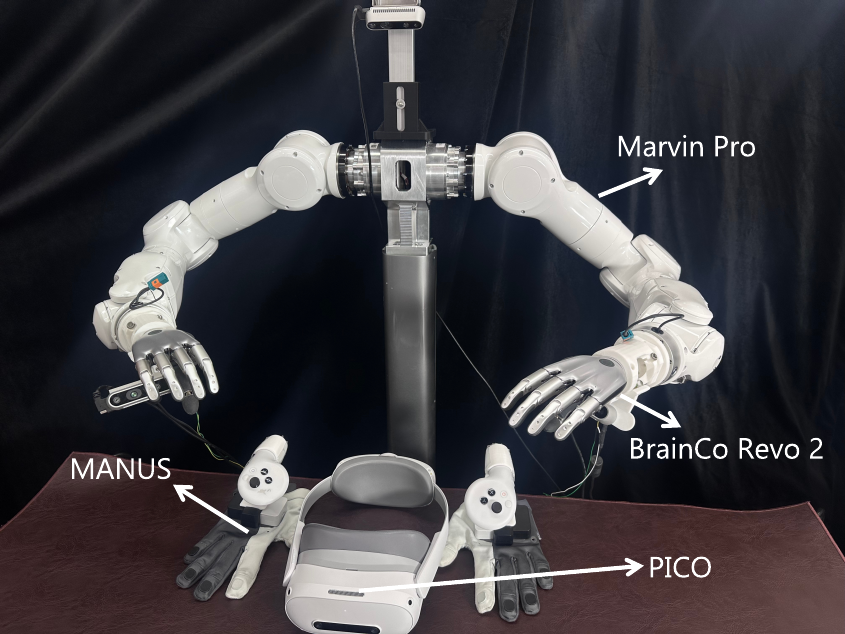}
    \caption{Real-world platform and teleoperation setup: Marvin Pro
    with two BrainCo Revo 2 hands, a PICO headset, and MANUS data gloves
    for demonstration collection.}
    \label{fig:real_robot_setup}
\end{figure}

\begin{figure*}[t]
    \centering
    \includegraphics[width=0.90\textwidth]{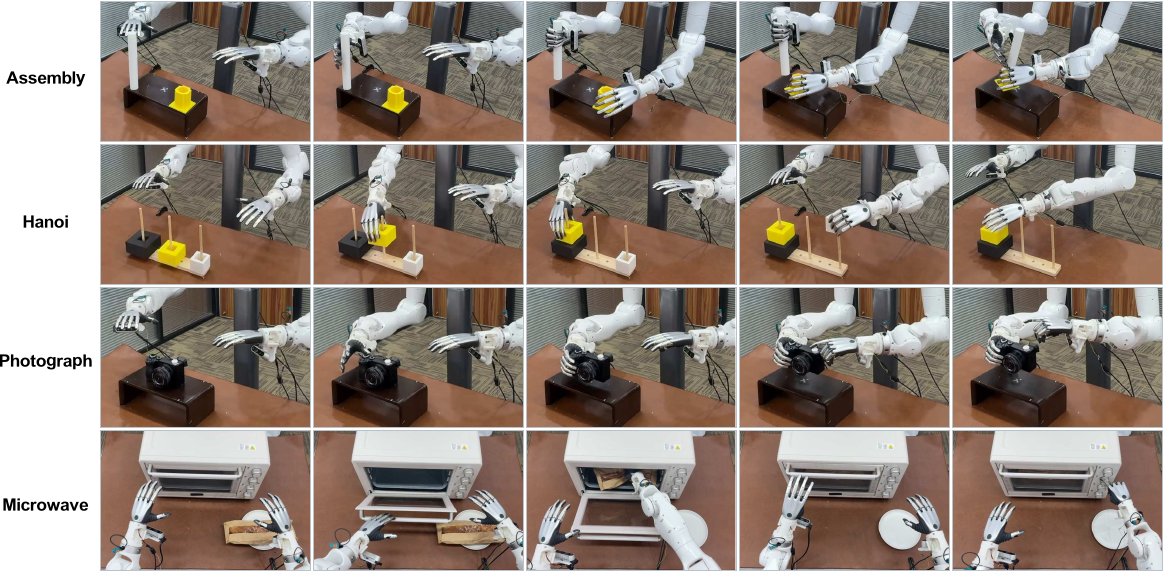}
    \caption{Representative real-world rollouts of Coordi-$\pi$.
    Each row shows one task, with successive stages ordered from left to right.}
    \label{fig:real_world_rollout}
\end{figure*}

\looseness=-1
As shown in Table~\ref{tab:ablation}, the interleaved action
representation alone achieves an average success rate of $39.7\%$.
Adding either the hand-identity embedding or factorized RoPE improves
the average performance, while combining both reaches $51.3\%$, an
$11.6$ percentage-point improvement over the interleaved baseline.
The larger gain of the full model shows that neither component alone
accounts for the improvement.

The largest gains over the interleaved baseline occur on
\textit{Unlock iPad}, \textit{Photograph}, and \textit{Hanoi}, all of
which involve sustained role differentiation or ordered cross-hand
interactions.

\looseness=-1
The effect of individual components is task-dependent. On
\textit{Microwave}, neither the hand-identity embedding nor
factorized RoPE alone improves over the interleaved baseline, whereas
their combination achieves the best performance. This non-monotonic
behavior also shows that the contribution of each component cannot be
understood independently of the full representation.
\Needspace{4\baselineskip}
\subsection{Bimanual Coordination Analysis}
\label{sec:bcs_analysis}

Success rate measures task completion, but does not characterize how the two arm--hand systems divide complementary
roles or satisfy task-level temporal dependencies during
execution. Prior work has studied role-structured bimanual
behavior and evaluation beyond binary task
success~\cite{buds,roboeval,bicoord}. We therefore evaluate
bimanual coordination using task-specific role and temporal
criteria, summarized by the Bimanual Coordination Score (BCS).

For each task, we define two complementary roles, $A$ and $B$, and let
$s(r,h)\in[0,1]$ denote the role evidence of hand $h$ for role $r$.
For manipulation or operation roles, $s(r,h)$ is the hand's share of
task-relevant contacts during the corresponding operation. For support
roles, it is the fraction of eligible support frames in which the hand
maintains stable contact with the corresponding object. We define the
Role Coordination Score (RCS) as
\begin{equation}
\mathrm{RCS}
=
50
\max_{h\in\{L,R\}}
\left[
s(A,h)+s(B,\bar{h})
\right],
\end{equation}
where $\bar{h}$ denotes the opposite hand. The maximization makes the score invariant to a global swap of
left- and right-hand role assignments.

For each prescribed temporal relation $k$, let $D_k$ denote the frames
of the dependent operation and $P_k$ the frames in which its
prerequisite is satisfied. The Temporal Coordination Score (TCS) is
\begin{equation}
\mathrm{TCS}
=
\frac{100}{K}
\sum_{k=1}^{K}
\frac{|D_k\cap P_k|}{|D_k|},
\end{equation}
where $K$ is the number of prescribed temporal relations. If a required
dependent operation is absent, its temporal term is set to zero. BCS is
reported as the equal-weighted average of RCS and TCS.

\textbf{Evaluation Details.}
BCS is evaluated under the \textit{rand-obj} setting using the same evaluation rollouts as the success-rate results. Role and temporal events are computed directly from simulator contacts, object poses, and task states using fixed task-specific rules shared by all evaluated policies. Role assignment is hand-agnostic, and a missing
required role contributes zero. The following note summarizes the
task-specific role and temporal criteria.

\par\smallskip
\begingroup
\setlength{\fboxsep}{4pt}
\setlength{\fboxrule}{0.5pt}
\noindent\fcolorbox{black!65}{black!3}{%
\begin{minipage}{\dimexpr\columnwidth-2\fboxsep-2\fboxrule\relax}
\small
\textbf{Task-specific BCS criteria.}
For support--operation tasks, the prerequisite requires stable support
by the non-operating hand.
\begin{list}{\textbullet}{%
  \setlength{\leftmargin}{1em}%
  \setlength{\labelsep}{0.4em}%
  \setlength{\topsep}{3pt}%
  \setlength{\itemsep}{1pt}%
  \setlength{\parsep}{0pt}%
}
\item \textbf{Hanoi:} Medium- and small-disk manipulation by opposite
hands; score small-disk manipulation while the medium disk remains
legally stable on the target post.

\item \textbf{Photograph:} Camera support and shutter operation;
score the first shutter-contact segment during stable support.

\item \textbf{Unlock iPad:} Tablet support and keypad operation;
score each of the first three keypad-contact segments during stable
support.

\item \textbf{Assembly:} Socket support and peg manipulation;
score peg--socket contact while the socket remains stably supported
off the table.

\item \textbf{Microwave:} Food manipulation and door operation;
score food contact with the door open, door closing while food is
inside, and the first start-button contact segment while food is
inside and the door is closed.
\end{list}
\end{minipage}%
}
\par
\endgroup
\smallskip

As shown in Table~\ref{tab:coordination}, Coordi-$\pi$ achieves the
highest average BCS across the five tasks and obtains the best BCS on
four of them. This indicates that the improvement in task success is
generally accompanied by more consistent satisfaction of the
task-specific role and temporal coordination criteria.

\textit{Hanoi} provides an informative counterexample. The interleaved
baseline achieves a higher BCS than Coordi-$\pi$ ($89.75$ vs.\
$86.32$), yet a lower success rate ($37.3\%$ vs.\ $48.7\%$). Thus,
satisfying the prescribed coordination pattern is not sufficient for
successful execution. A rollout may exhibit the intended division of
roles and temporal ordering while still failing due to execution errors
or task conditions not captured by BCS. This distinction is useful:
BCS measures the structure of bimanual coordination during execution,
whereas success rate additionally reflects whether that coordinated
behavior is executed accurately enough to complete the task.

\Needspace{5\baselineskip}
\subsection{Performance under Full Randomization}
\label{sec:randfull}
We further examine whether the advantage of Coordi-$\pi$ persists
under the stronger visual and environment variation of
\textit{rand-full}. Coordi-$\pi$ achieves an average success rate of
$32.3\%$, compared with $18.3\%$ for $\pi_{0.5}$, retaining a
$14.0$ percentage-point advantage under the more challenging setting.
The gains remain particularly clear on \textit{Unlock iPad} and
\textit{Hanoi}, indicating that the benefit of the structured action
representation persists even when perception becomes more variable.

At the same time, the absolute success rates are substantially lower
than under \textit{rand-obj}, and \textit{Assembly} remains particularly
challenging under \textit{rand-full}. Coordi-$\pi$ achieves a non-zero
success rate of $2.0\%$ on \textit{Assembly}, while $\pi_{0.5}$ records
no successful trials. This suggests that improving bimanual action
structure does not remove other bottlenecks associated with strong
visual variation and contact-sensitive execution. In particular, the
remaining failures are likely to depend on perception and precise
interaction in addition to the action representation itself.
\Needspace{4\baselineskip}
\subsection{Real-World Experiments}
\label{sec:real_world}

We further evaluate Coordi-$\pi$ on the TianJi Marvin Pro dual-arm
platform shown in Fig.~\ref{fig:real_robot_setup}, equipped with two
BrainCo Revo 2 dexterous hands, an external RGB camera, and
wrist-mounted cameras. Demonstrations are collected via teleoperation
using a PICO headset and MANUS data gloves. We consider four bimanual
dexterous manipulation tasks: \textit{Photograph}, \textit{Hanoi},
\textit{Assembly}, and \textit{Microwave}. Together, these tasks cover
object support and operation, sequential cross-hand interaction,
contact-rich assembly, and coordinated long-horizon execution.
Representative rollouts are shown in Fig.~\ref{fig:real_world_rollout}.

\suppressfloats[t]
\begin{figure}[t]
    \centering
    \includegraphics[width=\columnwidth,trim=0 8bp 0 7bp,clip]
    {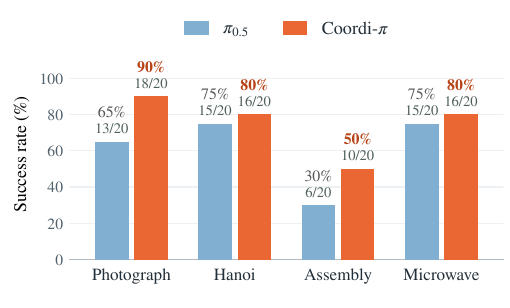}
    \caption{Real-world success rates of $\pi_{0.5}$ and Coordi-$\pi$.}
    \label{fig:real_world_results}
\end{figure}

For each task, we collect 100 demonstrations, resulting in 400
real-world demonstrations in total. Both $\pi_{0.5}$ and Coordi-$\pi$
are trained on the same demonstration set with the same training
budget. During evaluation, the two methods use the same initialization
range and task-success criteria. Each method is evaluated for 20 trials
per task, yielding 80 real-world trials per method.

As shown in Fig.~\ref{fig:real_world_results}, Coordi-$\pi$ succeeds in
60 of 80 trials, compared with 49 of 80 for $\pi_{0.5}$, increasing
the average success rate from $61.3\%$ to $75.0\%$. The improvement is
consistent across all four tasks, with the largest gains on
\textit{Photograph} and \textit{Assembly}. Together with the simulation
results, this consistent improvement shows that the advantage of the
proposed action representation is not limited to the simulated
evaluation setting.

The result on \textit{Assembly} is particularly informative. Although
the task remains difficult in simulation, Coordi-$\pi$ improves its
real-world success rate from $30\%$ to $50\%$. Since the simulation and
real-world evaluations differ in their data and execution conditions,
their absolute success rates are not directly comparable. Nevertheless, the improvement on \textit{Assembly} shows that the benefit of the
structured bimanual action representation is not limited to tasks with
comparatively higher success rates in simulation.

\Needspace{4\baselineskip}

\section{CONCLUSION}

We presented Coordi-$\pi$, a structured action representation for
bimanual dexterous manipulation that interleaves right- and left-side
arm--hand tokens, explicitly represents hand identity, and separates
sequence position from control timestep through factorized RoPE.
Across five DexJoCo tasks, Coordi-$\pi$ consistently improves over
$\pi_{0.5}$ under both \textit{rand-obj} and \textit{rand-full}, and
the same advantage is observed across four real-world tasks. The
ablation and coordination analyses further show that the gains are
most pronounced on tasks involving complementary hand roles and
ordered cross-hand interactions, and are generally accompanied by
better bimanual coordination during execution.

At the same time, precise contact-rich manipulation remains difficult,
particularly under stronger randomization. These results suggest that
structured bimanual action representation addresses an important
coordination bottleneck, while perception and precise interaction
remain complementary challenges for future work.

\bibliographystyle{IEEEtran}
\bibliography{references}

\end{document}